\documentclass[twoside,11pt]{article}

\usepackage{blindtext}

\usepackage{jmlr2e}

\usepackage{lastpage}
\jmlrheading{23}{2022}{1-\pageref{LastPage}}{1/21; Revised 5/22}{9/22}{21-0000}{}

\ShortHeadings{GEOMETRICAL GROUP ACTION AND CURIOSITY}{Sergeant, Ruet, Rudrauf, Ognibene and Tisserand}
\firstpageno{1}

 \usepackage{amsthm}
\usepackage{mathtools}
  \usepackage{dsfont}
 \usepackage{amsmath}
 \usepackage{amsthm}

\usepackage{comment}
\theoremstyle{definition}
 \newtheorem{defn}{Definition}
  
  \newtheorem*{defn*}{Definition}
 
  \theoremstyle{plain}
  
  \newtheorem{prop}{Proposition}
  \newtheorem*{prop*}{Proposition}
  \newtheorem{lem}{Lemma}
   \newtheorem*{cor*}{Corollary}
  \newtheorem*{theo*}{Theorem}
  \newtheorem*{thm*}{Theorem}
  
  \theoremstyle{remark}
  \newtheorem{rem}{Remark}
 
\usepackage[linesnumbered, ruled]{algorithm2e}
\usepackage{tikz}
\usepackage{tikz-cd}
\usetikzlibrary{positioning}
\DeclareMathOperator{\argmax}{argmax}
\DeclareMathOperator{\1}{1}

\begin{document}
\title{Influence of the Geometry of the world model on Curiosity Based Exploration}

 \author{\name Grégoire Sergeant-Perthuis \email gregoireserper@gmail.com \\ \addr LCQB Sorbonne Université \& OURAGAN team, Inria Paris Paris, France.
 \AND
 Nils Ruet\\
 \addr
CIAMS, Université Paris-Saclay, Orsay \& Université d'Orléans, Orléans, France.
 \AND
 David Rudrauf\\
 \addr
CIAMS, Université Paris-Saclay, Orsay \& Université d'Orléans, Orléans, France.
\AND
Dimitri Ognibene \\
\addr
 Dipartimento di Psicologia, Università Milano-Bicocca, Milan, Italy.
\AND
Yvain Tisserand\\
\addr
 CISA, University of Geneva, Geneva, Switzerland.
}


\editor{*}

\maketitle
\thispagestyle{empty}
\pagestyle{empty}


\begin{abstract}%
In human spatial awareness, 3-D projective geometry structures information integration and action planning through perspective taking within an internal representation space. The way different perspectives are related and transform a world model defines a specific perception
and imagination scheme. In mathematics, such collection of transformations corresponds to a ‘group’, whose ‘actions’ characterize the geometry of a space. Imbuing world models with a group structure may capture different agents’ spatial awareness and affordance schemes. We used group action as a special class of policies for perspective-dependent control. We explored how such geometric structure impacts agents’ behavior, comparing how the Euclidean \textit{versus} projective groups act on epistemic value in active inference, drive curiosity, and exploration behaviors. We formally demonstrate and simulate how the groups induce distinct behaviors in a simple search task. The projective group’s nonlinear magnification of information transformed epistemic value according to the choice of frame, generating behaviors of approach toward an object of interest. The projective group structure within the agent's world model contains the Projective Consciousness Model, which is know to capture key features of consciousness. On the other hand, the Euclidean group had no effect on epistemic value : no action was better than the initial idle state. In structuring \textit{a priori} an agent’s internal representation, we show how geometry can play a key role in information integration and action planning.

\end{abstract}

\section{Introduction}

In artificial agent learning and control, intrinsic and extrinsic rewards can be combined to optimize the balance between exploration and exploitation. 
Intrinsic rewards in Reinforcement Learning (RL) \citep{hester2017intrinsically, merckling2022exploratory, oudeyer2007intrinsic} or terms of epistemic value in active inference \citep{Friston2015} have been brought forth as mechanisms mimicking curiosity and driving exploration, e.g. by integrating prediction error or uncertainty to drive actions favoring their reduction. However, efficient exploration is a computationally hard task. Recent neural planning models have increased planning flexibility and generality \citep{sekar2020planning}. Yet, it is well-known that models' structures heavily impact planning performance  and tractability \citep{geffner2013concise} as well as learning complexity \citep{goyal2022inductive}. A good representation of information may improve learning and search efficiency. 

These issues are particularly salient for computation-heavy, highly recursive machine learning algorithms and applications, e.g. reinforcement learning (RL) in artificial agents \citep{bonet2019learning2} or recursive modeling method (RMM) in multi agent systems (MAS) and partially observable stochastic games (POSG) \citep{geffner2013concise}.
Although generic neural world models can support exploration-related processes, incorporating prior knowledge that shapes internal representations to more effectively support exploration across a broad range of environments, such as 3-D environments, may enable autonomous agents to explore more complex and realistic settings on a larger scale \citep{goyal2022inductive}. The exploration planning problem can thus be approached by considering how the structure of representation impacts exploration behaviors. In this article, the structure of representations is encoded into the geometry of the state space of an agent; we quantify the impact of changing this geometry on the behavior of the agent.

Here, we do not consider mechanisms of representation learning, e.g. in which world dynamics and action effects need to be learned and represented, as typically done in RL. We specifically consider control and execution when object locations, world states, and maps may not be known but dynamics, rewards, and action effects already are. We focus on action selection for environment exploration and mapping.

We adopt the active inference framework, i.e. an implementation of the Bayesian Brain Hypothesis aimed at generating adaptive behaviors in agents \citep{FRISTON200670}, that has found applications in neuroscience \citep{DACOSTA2020102447} and proposed for modeling molecular machines \citep{ijms222111868,timsit_evolution_2021}. It relies on an internal representation of the environment that an agent is driven to explore and exploit. The agent continually updates its beliefs about plausible competing internal hypotheses on the environment state. Under common sensory limitations, active inference relates to Partially Observable Markov Decision Process (POMDP) \citep{kaelbling1998planning,10.1007/978-3-030-35888-4_31}. The epistemic value of states is a quantity that arises in active inference \citep{Friston2015}. Its maximization drives the agent's curiosity and actions. 

For exploration or search in 3-D space, it is warranted to consider how geometrical principles could be embedded into efficient control mechanisms, to regularize the internal representation of information and mediate exploration under a drive of uncertainty reduction or information maximization. Geometrical considerations have previously been integrated into a variety of optimization and machine learning approaches, such as RL, active inference, and Bayesian inference (See Related Works below), but not in the specific perspective we introduce herein. 

We build upon the hypothesis that 3-D internal representations of space in agents performing active inference may correspond to specific geometries, with properties that can be exactly analysed. 
More specifically, we consider how different first person perspectives may relate to each other, through transformations of a world model, as a specific perception and imagination scheme for agents. This entails considering the action of geometrical groups of transformations (in the mathematical sense of the concept in Group theory; see Section \ref{group-def}) on the spatial distribution of information experienced and encoded by agents internally. The question is whether such group action could contribute to information gain estimation and maximization, as an internal planning or perspective-dependent control mechanism. Certain geometrical groups might imply internal representations, policies, value functions and principles of action that are particularly relevant for search and exploration. More specifically, we wish to compare how different groups impact the quantification of epistemic value. We then wish to characterize how the optimization of action from those different groups may yield different exploration behaviors. We contrasted two separate toy models of an agent performing a simple search task using active inference based solely on epistemic drives. One model used Euclidean geometry and the other projective geometry for the agent's internal space. We compared the two models in terms of resulting exploration behaviors and effects on epistemic value.\linebreak
\indent We chose to compare Euclidean versus projective geometry based on previous work, leveraging psychological research on the phenomenology of spatial consciousness and its role in the control of behaviors \citep{rudrauf2017mathematical,rudrauf2020moon,RUDRAUF2022110957,rudrauf2023combining}. This research suggested that 3-D projective geometry plays a central role in human cognition and decision-making by shaping information representation and subsequent drives. It also offers a mechanism of changes of points of view and perspectives on a world model, including for perspective taking in social cognition, which is critical for the development of strategies of action planning in humans (see  \citealt{RUDRAUF2022110957,rudrauf2023combining}). We used Euclidean geometry as a standard baseline geometry for comparison \citep{ognibene2013towards}.
Our geometrical rationale implies a different understanding of how agents' actions in their environment (here in the behavioral sense of the term) are implemented and selected compared to usual active inference. Agents' actions, such as navigation and approach-avoidance behaviors, can naturally be seen as dual to internal changes of perspective, i.e. group actions, in their representation space. We thus used group actions as a predictive model of actual behavioral actions.\linebreak
\indent The approach allowed us to formally study and demonstrate how the geometry governing the internal representation space may directly impact the computation of epistemic value and ensuing exploratory behaviors. Projective geometry \textit{versus} Euclidean geometry demonstrated remarkable properties of information integration for motion planning under epistemic drive.

\section{Related Works}

\subsection{Representation of Space and Exploration, in the Context of Machine Learning and Control}

The integration of geometrical mapping in machine learning has been proposed to reduce the high-dimensionality of input spaces and provide efficient solutions for action selection and navigation. Seminal neurally inspired models used projections on 2-D manifolds for representation learning of complex spatial information and self-motion effects \citep{1296691}. The impact of changes of perspective in exploration has long been of interest \citep{ognibene2013towards}. Ferreira et col. \citep{ferreira2013bayesian} proposed an internal 3-D egocentric, spherical representation of space, to modulate information sampling and uncertainty as a function of distance, and control a robot attention through Bayesian inference. This was a seminal example of how geometrical rationale could suggest solutions to integrate perception and action planning.

Exploration methods must often maintain high-resolution representations of space to maximize information gain following action. This may hinder exploration efficiency, in particular in large-scale environments. 3-D topological representations of ambient space have been proposed as part of an abstract planning scheme, showing promising improvements of exploration efficiency \citep{9561830}. 

Active vision principles, combined with curiosity-based algorithms and RL, were applied to the learning of saliency maps in the context of autonomous robots' navigation \citep{7759716}. The approach yielded promising optimization solutions to both adaptive learning of task-independent, spatial representations, and efficient exploration policies, which could serve as prior to support long-term, task-oriented, utility-driven RL mechanisms \citep{7759716} (see also \citealp{ognibene2014ecological,sperati2017bio}).\\
\indent Complex control tasks with continuous state and action spaces have been solved using deep reinforcement learning (DRL) with joint learning of representations and predictions. Such approach may entail non-stationarity, risks of instability and slow convergence, in particular in control tasks with active vision. Separating representation learning and policies' computations may mitigate the issues, but may also lead to inefficient information representations. Merckling et col. \citep{merckling2022exploratory} have sought to build compact and meaningful representations based on task-agnostic and reward-free agent-environment interactions. They used (recursive) state representation learning (SRL) while jointly learning a state transition estimator with near-future prediction objective, to contextually remove distracting information and reduce the exploration problem complexity. Positive outcomes were maximized through inverse predictive modeling, and prediction error was used to favor actions reducing uncertainty, which improved subsequent performance in RL tasks. The authors emphasized that dealing with partial observability through memory and active vision may require new solutions to both representation learning of hidden information and exploration strategies.\\
\indent Uncertainty-based methods using intrinsic reward and exploration bonuses to plan trajectories have been criticized for inducing non-stationary decaying surprise, and for being hard to structure and optimize \citep{guo2021geometric}. Maximum State-Visitation Entropy (MSVE) was introduced to maximize state exploration uniformity, but optimization has been often challenging for large state spaces. Guo et col. \citep{guo2021geometric} have introduced Geometric Entropy Maximization, which leverages geometry-aware entropy based on Adjacency Regularization (AR) and a similarity function, in order to optimize the MSVE problem at scale. \\
\indent Geometrical constraints considered across these related works were not integrated into a global model, and were somewhat \textit{ad hoc}. They pertained to a lower level of processing than the one we are concerned with here. However, they emphasize the current needs and challenges for integrating geometry in learning, control and navigation.\\
\indent Methods and algorithms combining computer vision, machine learning and optimization, e.g. for robotic planning, have integrated group theoretic concepts to obtain, for instance, invariance to rotation and translation in image processing \citep{lee2004geometric, qin2019autonomous, meng2017intelligent}. Likewise, the leveraging of geometrical concepts, in Deep learning, e.g. for learning manifolds and graphs, has been growing in recent years, demonstrating very promising results for representation learning \citep{gerken2021geometric, cao2022geometric}. The approach introduces combinatorial structures to leverage prior knowledge of geometry on the data of interest, e.g. applying `convolutional Neural Networks' to non-Euclidean space. However, the Euclidean group $E_3$, or more specifically $SE_3$ (see \citealp{lee2004geometric}), which includes translations and rotations, but excludes reflections, or simply $SO_3$, the 3-dimensional rotation group \citep{gerken2021geometric}, are the groups being typically considered. \\
\indent Here, in addition to the Euclidean group, we also consider $PGL(\mathbb{R}^3)$, the projective general linear group in 3-D, which acts on a projective space through projective transformations. The projective group is central to computer vision, for instance to generate 2-D images from 3-D information, but is used in such context in a restricted manner. We sought approaches based on cognitive science, considering spatial cognition and its relations to action at a higher level of integration, which does not reduce to the visual modality, but instead assume the mapping of multimodal information on a supramodal internal space of representation.   

\subsection{Projective Consciousness Model (PCM) and Active Inference}

It has been shown that geometrically constrained active inference can be used as a framework to understand and model central aspects of human spatial consciousness, through the Projective Consciousness Model (PCM) \citep{rudrauf2017mathematical, RUDRAUF2022110957}. According to this model, consciousness accesses and represents multimodal information through a Global Workspace \citep{dehaene2017consciousness} within which subjective perspectives on an internal world model can be taken. The process contributes to appraise possible actions based on their expected utility and epistemic value \citep{RUDRAUF2022110957}. In publications on PCM \citep{rudrauf2017mathematical, williford2018projective,RUDRAUF2022110957,rudrauf2023combining,williford2022pre}, it was hypothesized that such internal representation space is geometrically structured as a $3-D$ projective space, denoted $P_3(\mathbb{R})$. Changes of perspective then correspond to the choice of a projective transformation $\psi$, i.e. an action from $PGL_3$. A projective transformation can also be modeled as a linear isomorphism $M_\psi \in GL_4(\mathbb{R})$ up to a multiplicative constant. The model yieled an explanation for the Moon illusion \citep{rudrauf2020moon} with falsifiable predictions on how strong the effect should be depending on context; as well as for the generation of adaptive and maladaptive behaviors, consistent with developmental and clinical psychology (see \citealp{RUDRAUF2022110957}). Though essential in integrative spatial cognition, notably for understanding multi-agent social interactions, perspective taking is rarely integral to existing models of consciousness or formally implemented \citep{Tononi,10.3389/fams.2020.602973,MASHOUR2020776,dehaene2017consciousness,merker2022integrated}. The PCM assumes that projective mechanisms of perspective changes are integral to the global workspace of consciousness, both in non-social and social contexts. The advantages of mechanisms of perspective taking for cybernetics remains to be fully formulated (see \citealp{RUDRAUF2022110957}).

\section{Methodology}

The experiment we considered is that of an agent, denoted as $a$, which is looking for an object $O$ in the `real world', the $3$-D Euclidean space $E_3:=\mathbb{R}^3$. The set of moves of the agent is denoted $M$. The position of $O$ is denoted $o\in E_3$. 
The agent `represents' the position of the object $O$ inside its `internal world model'. We consider `internal world models', spaces denoted $W$, that are such that there is a group acting on them; we call such spaces, group structured world models. This group accounts for the change of coordinates that each movement of the agent induces when the positions of the object are expressed in the agents reference frame.  We consider two spaces in particular:
\begin{enumerate}
    \item \emph{Euclidean case}: $W$ is the 3-D vector space, $W= \mathbb{R}^3$
    \item \emph{Projective case}: $W$ is the 3-D projective space, denoted as $P_3(\mathbb{R})$
\end{enumerate}

We will denote $\mathcal{B}(W)$ the Borel $\sigma$-algebra of the respective topological spaces. 

\begin{figure}
    \centering
    \includegraphics[width=0.4\textwidth]{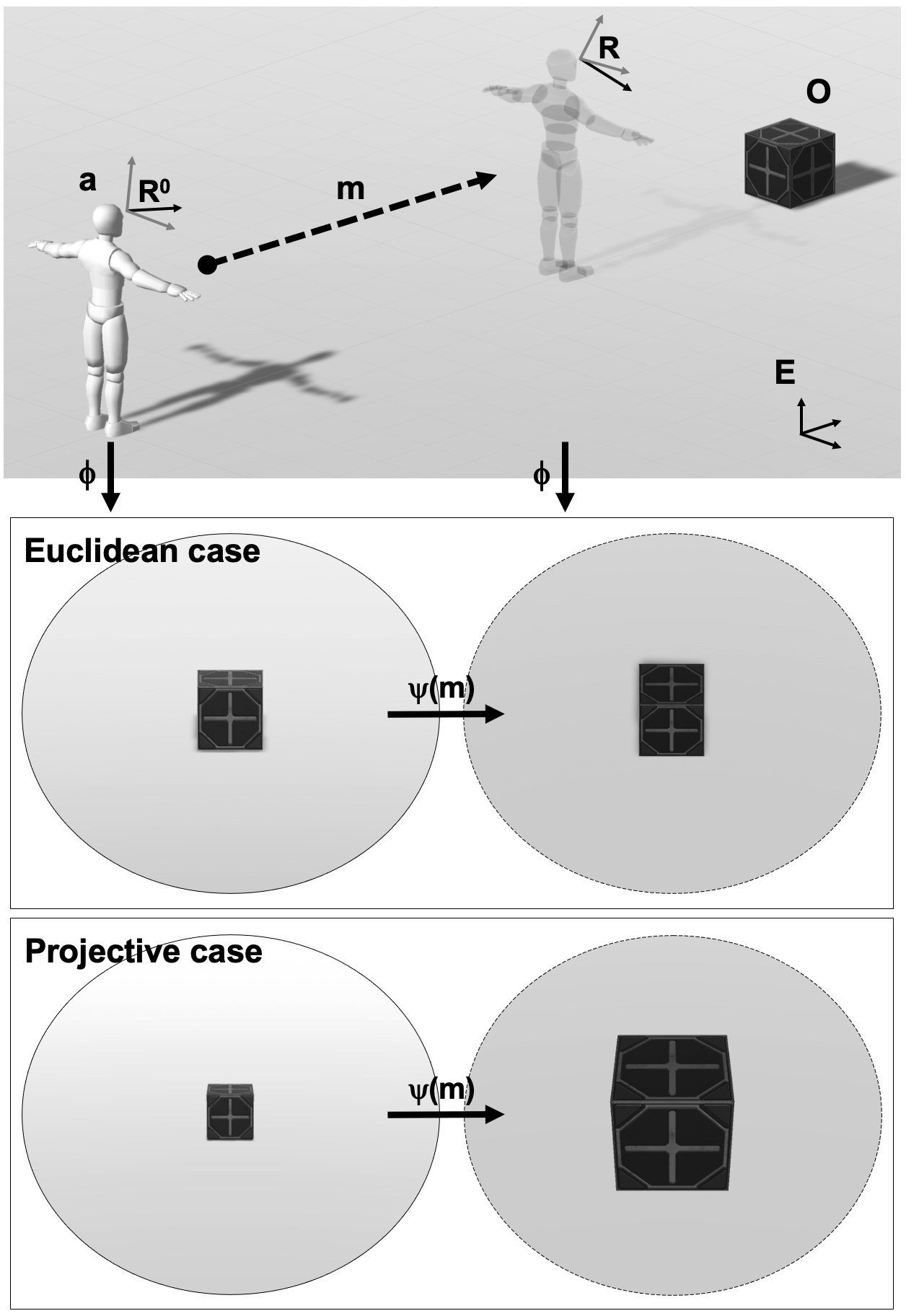}
    \caption{Toy model setup and main transformations}
    \medskip
    \small
    \raggedright
    {\textit{Upper-tier.} Agent $a$ simulates move $m$ in Euclidean space $E$. $\mathcal{R}^0$ and $\mathcal{R}$ are its frames in $E$ before and after the move, oriented toward object $O$. Vertical arrows indicate transformations $\phi$ from the external to the internal space. \textit{Lower-tier.} Rendering of the effect of the internal group action $\psi(m)$ corresponding to move $m$ in the Euclidean \textit{versus} projective case. (Made with Unity).}
    \label{fig:figure_01}
\end{figure}

In Section \ref{relating}, we explain how the `real world' and the `internal world model' are related to one another in both the Euclidean and Projective case. Figure \ref{fig:figure_01} illustrates the setup of the toy model and main transformations considered. The agent's internal beliefs about the position of the object are encoded by a probability measure on $W$ that the agent updates through observations. The agent explores its environment through the computation of an epistemic value, the maximization of which captures curiosity-based exploration. In Section \ref{epistemic value}, we explain how epistemic value is defined for group structured internal representations. In Section \ref{algorithm} we give the details of the exploration algorithm. 

\subsection{Group Structured World Model}\label{group-def}

Let us first recall what a group is. 
\begin{defn}[Group, §2 Chapter 1 \citealp{lang2012algebra}]
A group is a set $G$ with an operation $.: G\times G\to G $ that is associative, such that there is an element $e\in G$ for which $e.g=g$ for any $g\in G$, and any $g\in G$ has an inverse denoted $g^{-1}$ defined as satisfying, $g.g^{-1}= g^{-1}.g=e$.
\end{defn}

We call a group structured world model, a world model provided with a group action; we now make this statement formal. 
\begin{defn}[Group structured world model]
$W$ is a group structured world model for the group $G$ when there is a map $h: G\times W\to W$ denoted as $h(g,x)=g.x$ for $g\in G$ and $x\in X$, such that, 

\begin{enumerate}
    \item $(g.g_1).x= g.(g_1.x)$ for all $g,g_1\in G$, $x\in W$
    \item $e.x=x$, for all $x\in W$
\end{enumerate}
\end{defn}

In the \emph{Euclidean case} the group structured world model, $W$, is the 3-D vector space $\mathbb{R}^3$; it is structured by the group of invertible matrices $GL_3(\mathbb{R})$. In the \emph{Projective case}, the group structured world model, $W$, is the projective space $P_3(\mathbb{R})$; it is structured by the group of projective linear transformations $PGL(\mathbb{R}^3)$.

\subsection{Relating the `Real World' to the `Internal World Model'}\label{relating}

We assume that the `real world' is the 3-D Euclidean space, $E_3$. 
We assume that the `real world' comes with with an Euclidean frame $\mathcal{R}_E$, i.e. a point $\mathcal{C}$ and three independent vectors $e_0,e_1,e_2$. This frame is used to set up the experiment: the configurations of the object and agent across time are encoded in this frame; it is fixed once and for all before starting the experiment. Therefore we now identify $E_3$ with $\mathbb{R}^3$, $\mathcal{C}$ with $(0,0,0)$ and $e_0,e_1,e_2$ with the respective basis vectors, $(1,0,0),(0,1,0),(0,0,1)$. The agent, denoted as $a$, is modeled as a solid in the `real world'; it has its own Euclidean frame (the solid reference frame), $\mathcal{R}:=(\mathcal{P},u_{0},u_{1},u_{2})$, with $\mathcal{P}$ the center of $a$ and $u_{0},u_{1},u_{2}$ three unitary vectors that form a basis. 

In the \emph{Euclidean case}, the map that relates $E_3$ and its group structured world model, $W$, is the affine map, $\phi_{\mathcal{R}}$, that changes the coordinate in $\mathcal{R}_E$ to coordinates in $\mathcal{R}$. 

In the \emph{Projective case}, this map is a projective transformation. The choice of such a projective transformation is dictated by Proposition A.1 \citep{RUDRAUF2022110957}. Let us now recall some facts about that transformation.

Let for any $(x,y,z)\in \mathbb{R}^3$,
\begin{equation}
\rho(x,y,z)= \left(\frac{x}{\gamma z+1}, \frac{y}{\gamma z+1}, \frac{z}{\gamma z+1}\right)
\end{equation}

with $\gamma\in \mathbb{R}_{+}$ a strictly positive parameter. 

The (projective) transformation $\phi^p_{\mathcal{R}}$, from $E_3$ to $W$, which relates the `real world' to the `internal world model' in the projective case, is posed to be $\phi^p_{\mathcal{R}}:=\rho \circ \phi_{\mathcal{R}}$.

\begin{prop}\label{change-frame}
When the agent $a$ makes the move $m\in M$, its solid reference frame changes from $\mathcal{R}$ to $\mathcal{R}^m$. In the \emph{Euclidean case} this move induces an invertible affine map, from the `internal world model' to itself. In the \emph{Projective case} it induces a projective transformation, $\psi_m\in PGL(\mathbb{R}^3)$.
\end{prop}

We denote $\1_U$ or $x\to \1[x\in U]$ the characteristic function of subset $U$, i.e. the functions that is equal to $1$ for $x\in U$ and $0$ elsewhere. 
\begin{rem}
In both cases there is a dense open subset, $U$, of $W$ which is in continuous bijection with $\mathbb{R}^3$. From the Lebesgue measure $dx$ on $\mathbb{R}^3$, we define the following measure on $W$, $d\lambda :=\1_U dx$. In what follows we do not raise this technical point anymore and simply refer to $d\lambda$ as the Lebesgue measure on $W$. 
\end{rem}

\subsection{Beliefs, Policies and Epistemic Value}\label{epistemic value}

\subsubsection{Beliefs} 

The agent $a$ keeps internal beliefs about the position of the object represented in its `internal world model'; these beliefs are encoded by a probability measure $Q_X\in \mathbb{P}(W)$, where $\mathbb{P}(W)$ denotes the set of probability measures on $W$. Probability measures will be denoted with upper case letters and their densities with lower case letters. These beliefs are updated according to noisy sensory observations of the position of $O$. `Markov Kernels' can be used to formalize noisy sensors. Let us recall their definition. 

A `Markov Kernel' $\Pi$ from $\Omega_1$ to $\Omega$ is a map $\Pi:\Omega\times \Omega_1\to [0,1]$ such that for any $\omega_1\in \Omega_1$, $\sum_{\omega\in \Omega} \Pi(\omega\vert \omega_1)=1$, i.e. a map that sends any $\omega_1\in \Omega_1$ to a probability measure $\Pi_{\vert \omega_1}\in \mathbb{P}(\Omega)$.

The uncertainty on the sensors of $a$ is captured by a Markov kernel $P_{Y\vert X}$ from $W$ to $W$. It is a parameter of the experiment: it is fixed before the agent starts looking for $O$. The couple $(P_{Y\vert X},Q_X)$ defines the following probability density, $P_{X,Y}\in \mathbb{P}(W\times W)$: for any $x,y\in W$,

\begin{equation}
P_{X,Y}(dx,dy):= p_{Y\vert X}(y\vert x) q_X(x)dxdy
\end{equation}

where $dx$ is the Lebesgue measure on $W$. An observation of the position of the object $y^{o}\in W$ triggers an update of the belief $Q_X$ to the belief with density

\begin{equation}
Q_{X\vert y^{o}}= \frac{p_{X,Y}(x\vert y^{o}) q_X(x) dx}{\int_{x\in W} p_{X,Y}(x\vert y^{o}) q_X(x) dx}
\end{equation}

\subsubsection{Policies} Recall that the agent has a set of moves it can make $M$; moves $m\in M$ are associated with the group action $\psi_m: W\to W$ (Proposition \ref{change-frame}). The agent plans the consequence of its moves on its internal world model one step ahead: each change of frame induces the following Markov Kernel, for any $m\in M$, $A\in \mathcal{B}(W)$, and $x_0\in W$,

\begin{equation}
p_{X_1\vert X_0,m}(A\vert x_0, m)= \1[\psi_m(x_0)\in  A]
\end{equation}

Each move $m$ spreads a prior $Q_X$ on $X_0$ into the following prior on $X_1$: $\forall A\in \mathcal{B}(W)$,

\begin{align}
\psi_{m,*}Q_X(A)&:= \int \1[\psi_m(x_0)\in  A] q_{X}(x_0) dx\\
 & = Q_X(\psi_m^{-1}A)
\end{align}

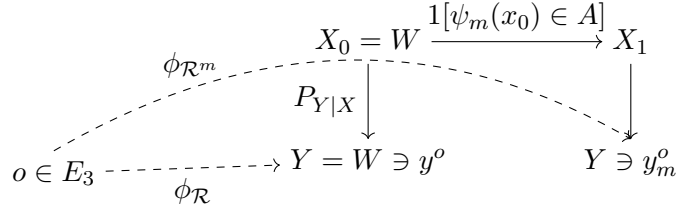
\begin{figure}[!h]
\centering

\begin{tikzpicture}


\node      (X_0)             {$X_0=W$};
\node      (E_0)     [left=of X_0]   [anchor=north east]{$\phi_{\mathcal{R}^m}$}  ;
\node      (E_2)     [below=of E_0]  {} ; \node      (E_1)      [left=of E_2]     {$o\in E_3$};
\node     (X_1)       [right=of X_0] {};
\node     (X_2)       [right=of X_1] {$X_1$};
\node     (Y_0)       [below=of X_0] {$Y=W 	\ni y^{o} $};
\node     (Y_1)       [below=of X_2] {$Y \ni y_m^{o}$};

\draw[->] (X_0) -- (X_2)   
node[midway,sloped,above] {$1[\psi_m(x_0)\in  A]$};
\draw[->] (X_0) -- (Y_0)  node[midway, left] {$P_{Y\vert X}$};
\draw[->] (X_2) -- (Y_1);
\draw[->,dashed] (E_1) -- (Y_0) node[midway, below]{$\phi_{\mathcal{R}}$};
\draw[->,dashed] (E_1.north) to [out=30,in=150] (Y_1.north);
\end{tikzpicture}

\caption{$m\in M$ is a move of the agent $a$, $1[\psi_m(x_0)\in  A]$ defines the kernel induced by move $m$, $P_{Y\vert X}$ is the noisy sensor. The diagram constituted of solid arrows defines the generative model the agent uses to plan its actions. $o$ is the position of the object in the `real world', $y^{o}\in W$ is the representation of $o$ in the `internal world model' of $a$ with respect to the solid reference frame $\mathcal{R}$, $y^{o}_m$ is the same for the reference frame $\mathcal{R}^m$ after move $m$.}
\label{figure:generative-model}
\end{figure}

We chose to denote this probability measure as $\psi_{m,*}Q_X$, because it is the standard mathematical notation for the `pushforward measure' by $\psi_{m}$. The generative model the agent uses to plan its future actions is summarized in Figure \ref{figure:generative-model}.

\subsubsection{Epistemic Value} 

The following definition is a restatement of the \emph{epistemic value} introduced in \citep{Friston2015} in the case of the kernel $P_{Y\vert X}:W\to W$. 

\begin{defn}[Epistemic Value]
For any probability measure $Q_X\in \mathbb{P}(W)$, the epistemic value of this measure is: 
\begin{align}\label{epistemic-defn}
C(Q_X):=&\mathbb{E}_{P_{Y}}\left[ H(P_{X\vert Y} \vert Q_X)\right]\\
&=\int p_{Y}(y)dy \int p_{X\vert Y}(x\vert y) \ln \frac{p_{X\vert Y}(x\vert y)}{q_X(x)} dx
\end{align}

$H$ is the relative entropy, also called Kullback-Leibler divergence.
\end{defn}

Reexpressing Equation \ref{epistemic-defn}, it becomes apparent that epistemic value is simply a mutual information: 

\begin{equation}\label{curiosity}
C(Q_X)=\int p_{X,Y} \ln \frac{p_{X,Y}(x,y)}{p_Y(y)q_X(x)} dx dy
\end{equation}

We propose to define the epistemic value of move $m$ as the epistemic value of the induced prior on $X_1$,

\begin{equation}
C(m):= C(\psi_{m,*}Q_X)
\end{equation}

\subsection{Exploration Algorithm}\label{algorithm}

Let us now put the previous elements together to describe the exploration behavior programmed in our agent. 
The agent $a$ is initialized in a configuration of the `real world', with solid reference frame $\mathcal{R}^0$; the object $O$ is positioned at $o\in E_3$. $a$ starts with an initial belief $Q_{X}^0\in \mathbb{P}(W)$ on the position of $O$. It plans one step ahead the consequence of move $m$; move $m$ induces a group action $\psi_m:W\to W$ that pushes forward the belief $Q_{X}^0$ to $\psi_{m,*}Q_{X}^0$. The agent then evaluates the epistemic value of $(P_{Y\vert X},\psi_{m,*}Q_{X}^0)$ for each move $m$ and chooses the move that maximizes this value, $m^{*}$. $a$ executes the move $m^{*}$ which transforms its solid reference frame $\mathcal{R}_0$ to $\mathcal{R}$. It can then observe (with its 'noisy sensors') the position of $O$ which is $y^{o}:=\phi_{\mathcal{R}}(o)$ in its internal world model, which triggers the update of prior $\psi_{m,*}Q_{X}^0$ to the distribution conditioned on the observation: $\left(\psi_{m,*}Q_{X}^{0}\right)_{\vert y^{o}_m}$. The process is iterated with this new prior. The exploration algorithm is summarized in Algorithm 1.

\begin{algorithm}[hbtp]\label{exploration-algorithm}

\KwData{{\small Initialization: $Q_{X}^0$ initial belief, $\mathcal{R}^0$ initial solid reference frame of $a$} }

$Q_X\gets Q_{X}^0$\;
\While{True}{

$m^* \gets \argmax_{m\in M} C(\psi_{m,*} Q_X)$\;
$\mathcal{R} \gets \text{solid reference frame of } a \text{ after move } m^*$\;
$Q_X\gets \psi_{m,*} Q_X$\;
$y^{o}\gets \phi_{\mathcal{R}}(o)$\;
$Q_X\gets Q_{X\vert y^{o}}$\;
}

\caption{{\small Curiosity based Exploration for agent} $a$}
\end{algorithm}

\section{Theoretical Predictions}

We wish to understand how the group by which the internal world model is structured influences the exploration behavior of the agent. The \emph{Euclidean case} serves as the reference model; in this case the world model shares the same structure as the real world: it is the `classical' way of modeling this exploration problem. The \emph{Projective case} corresponds to the hypothesis underlying the PCM. The following Theorem states that this experiment allows us to discriminate when the behavior of the agent is dictated by `objective' perspectives (Euclidean change of frame) \textit{versus} by `subjective' perspectives (projective change of frame) on its environment.

We consider the following noisy sensor, for any $x,y\in \mathbb{R}^3$,

\begin{equation}\label{noisy-sensor-kernel}
P_{Y\vert X}(y\vert x)=  \frac{3}{4\pi \epsilon^3} \1[\Vert x-y\Vert \leq \epsilon]
\end{equation}

where $\Vert . \Vert $ designates the Euclidean norm on $\mathbb{R}^3$, i.e. $\Vert x\Vert^2 = x_0^2+x_1^2+x_2^2$; $\epsilon >0$ is a strictly positive real number.

\begin{thm*}[Discrimination of behavior with respect to internal representations]\label{thm}

Let us assume that staying still is always a possible move for the agent. 

\emph{Euclidean case:} when the agent has an objective representation of its environment, given by an affine map, the agent stays still. \\ 

\emph{Projective case:} Assume now that the set of moves $M$ is finite; assume furthermore that after any possible move, the agent faces $O$, in other words, we assume that the agent knows in which direction to look in order to find the object but is still uncertain on \emph{where} the object is exactly. If it has a `subjective' perspectives, i.e. its representation is given through a projective transformation, it will choose the moves that allows it to approach $O$ (for any $\epsilon$ small enough). 

\end{thm*}

\begin{proof}
The details of the proof are given in Appendix \ref{appendix-1}. Let us here sketch the proof. The agent circumscribes a region of space in which it believes it is likely to find the object. This region corresponds to the error the agent tolerates on the measurement it makes of the position of $O$; we can also see it as the precision up to which the agent measures the position of $O$. In the \emph{Euclidean case}, the region in which the agent circumscribes the object appears to always be of the same size, irrespective of the agent's configuration with respect to the object. Therefore not moving ends up being an optimal option and the agent will not approach the object without additional extrinsic reward. In the \emph{Projective case}, the agent can `zoom' on this region in order to gain more precision in measuring $o$; the configurations of the agent in which this region is magnified are more informative regarding the position of $O$ and therefore preferred by the agent. The only way for the agent to actually zoom onto this area is to approach the location it believes $O$ is likely to be, therefore the agent will end up approaching $O$.

\end{proof}

\begin{rem}
This particular choice of Markov kernel (Equation \ref{noisy-sensor-kernel}) allows for an explicit expression of epistemic value which simplifies the proof of the result; however we expect the result to hold for a larger class of kernels. 
\end{rem}

In the next section we present an implementation of this experiment and simulation results.

\section{Simulations}

\subsection{Methods}

 Algorithm \ref{exploration-algorithm} is implemented in the following manner (source code available at \url{https://github.com/NilsRuet/effect-of-geometry-on-exploration}). Beliefs and the Markov kernel corresponding to sensors were considered to be multivariate normal distributions, that is $P_{Y\vert X} \sim \mathcal{N}(\mu_{Y \vert X},\,\Sigma_{Y \vert X})$ and $Q_X \sim \mathcal{N}(\mu_X,\,\Sigma_X)$. Belief update through the action of a group was approximated using a Gaussian distribution; a projective transformation changes a Gaussian distribution into a non Gaussian one which is difficult to describe. Therefore we replace this non-Gaussian distribution with a Gaussian distribution with same mean and variance. 

We assumed $\mu_{Y \vert X} = x$ (which implies $\mu_y=\mu_x$) and $\Sigma_{Y \vert X} = \epsilon^2\mathbb{I}$ where $\mathbb{I}$ is the identity matrix and $\epsilon>0$ a positive real number. As a result,  for a given observation $y^o$, $Q_{X\vert{y^o}}$ and $C(\psi_{m,*} Q_X)$ can be computed efficiently. The joint distribution $P$ on $X, Y$ is a Gaussian distribution:
\[P(x,y) =  P_{Y\vert X}(y\vert{x})Q_X(x)\]
\[P(x,y) \sim \mathcal{N}(\mu_{X,Y},\,\Sigma_{X,Y})\]
with
$\mu_{X,Y} = (\mu_X, \mu_X)$ and 
\begin{equation}
\Sigma_{X,Y} = \begin{pmatrix}
  \Sigma_{XX} &  \Sigma_{XX}\\ 
   \Sigma_{XX}&  \epsilon^2\mathbb{I}+\Sigma_{XX}
\end{pmatrix}
\end{equation}

The variance of $Y$ is $\Sigma_{YY} = \epsilon^2\mathbb{I}+\Sigma_{XX}$.

The joint distribution being Gaussian entails that the distribution of $X$ conditioned on $y=y^o$ is also Gaussian, thus $Q_{X\vert{y^o}} \sim \mathcal{N}(\mu_{X\vert{y^o}},\,\Sigma_{X\vert{y^o}})$. Applying Proposition 3.13.\citep{10.2307/20461449} to our setting, the mean and covariance of the conditioned distribution are given by:
\begin{equation}
\mu_{X\vert{y^o}} = \mu_X + \Sigma_{XX} \Sigma_{YY}^{-1} (y^o - \mu_X)
\end{equation}
\begin{equation}
\Sigma_{X\vert{y^o}} = \Sigma_{XX} - \Sigma_{XX}\Sigma_{YY}^{-1}\Sigma_{XX}
\end{equation}

Epistemic value is computed using the Kullback-Leibler divergence. With full knowledge of the joint distribution, in the Gaussian case, following the expression of entropy for gaussian vectors (Chapter 12 Equation (12.39) \citealp{10.5555/1146355}) it is computed as:
\begin{equation}
C(Q_X) = I(X; Y) = \frac{1}{2} \ln{\frac{ (\det{\Sigma_{XX}})(\det{\Sigma_{YY}}) }{\det{\Sigma_{X,Y}}}}
\end{equation}

The set of moves that can be selected by the agent is restricted to translations as the agent must always face the object. 
This constraint, as well as the choice of a simple model of noisy sensor (with homogeneous precision and resolution), was motivated by the aim of making formal demonstrations of theorems tractable, and implementations tightly related to the theoretical predictions. This is a departure from how spatial sampling typically operates in perception, e.g. decreasing resolution of the visual field with eccentricity (see \citep{RUDRAUF2022110957,rudrauf2023combining} for more realistic but less formally analyzable applications of a broader version of the PCM). However, the specific aim herein of demonstrating fundamental properties of the action of different geometrical groups on epistemic value and ensuing behaviors of approach motivated such restrictions. 

The set of possible translations is composed of eight translations with the same norm, with evenly distributed angles (one of them being oriented toward the object irrespective of the position of the agent), and also contained an idle state, i.e. no translation. Here the angles correspond to the angles of the translation and not a rotation angle of the solid frame of the agent as the agent must always face the object. 

We approximated the belief after the action $m$ of a given group using a Gaussian distribution, $\psi_{m,*} Q_X \sim \mathcal{N}(\mu_{m},\,\Sigma_{m})$. The mean and covariance matrix are approximated using numerical integration:

\begin{equation}
\mu_m = \int{xp(\psi_m^{-1}(x)){\frac{1}{\vert  \det J_{\psi_m}(\psi_m^{-1}(x)) \vert}}dx}
\end{equation}

\begin{equation}
\Sigma_m = \int{(x - \mu_m)(x - \mu_m)^{T}p(\psi_m^{-1}(x)){\frac{1}{\vert \det J_{\psi_m}(\psi_m^{-1}(x)) \vert}}dx}
\end{equation}

We ran two sets of simulations. In the first one (Figure \ref{fig:simus} left tier), the agent started from an initial position with the object always located at a fixed position, and the algorithm was applied for 20 iterations, for both the Euclidean and Projective internal spaces. The agent started at $(0,0)$ and the object was located at $(0,2)$ in the world frame $E_2$ spanning the agent's displacement floor. If all translations were associated with epistemic values that only varied within a small range ($\pm1e-4$) as compared to the epistemic value of the idle state, reflecting numerical imprecision, the idle state was selected (the agent did not move). The aim of this set of simulations was to compare trajectories of agents displacements through time across the two geometries. In the second set of simulations (Figure \ref{fig:simus} right tier), the agent started at the center of a $20\times 20$ grid of possible positions of the object. 

The positions that are too close to the agent are excluded so that a sufficient effect can be observed. For each object position, we considered only the first set of translations that the agent could envision from its initial position. We computed epistemic value across the set of possible translations of the agent for each object position and for both the Euclidean and projective internal spaces. The aim of this set of simulations was to be able to systematically compare epistemic values across the two geometries for all possible positions of objects.    

\subsection{Results}

\begin{figure}
    \centering
    \includegraphics[width=0.8\textwidth]{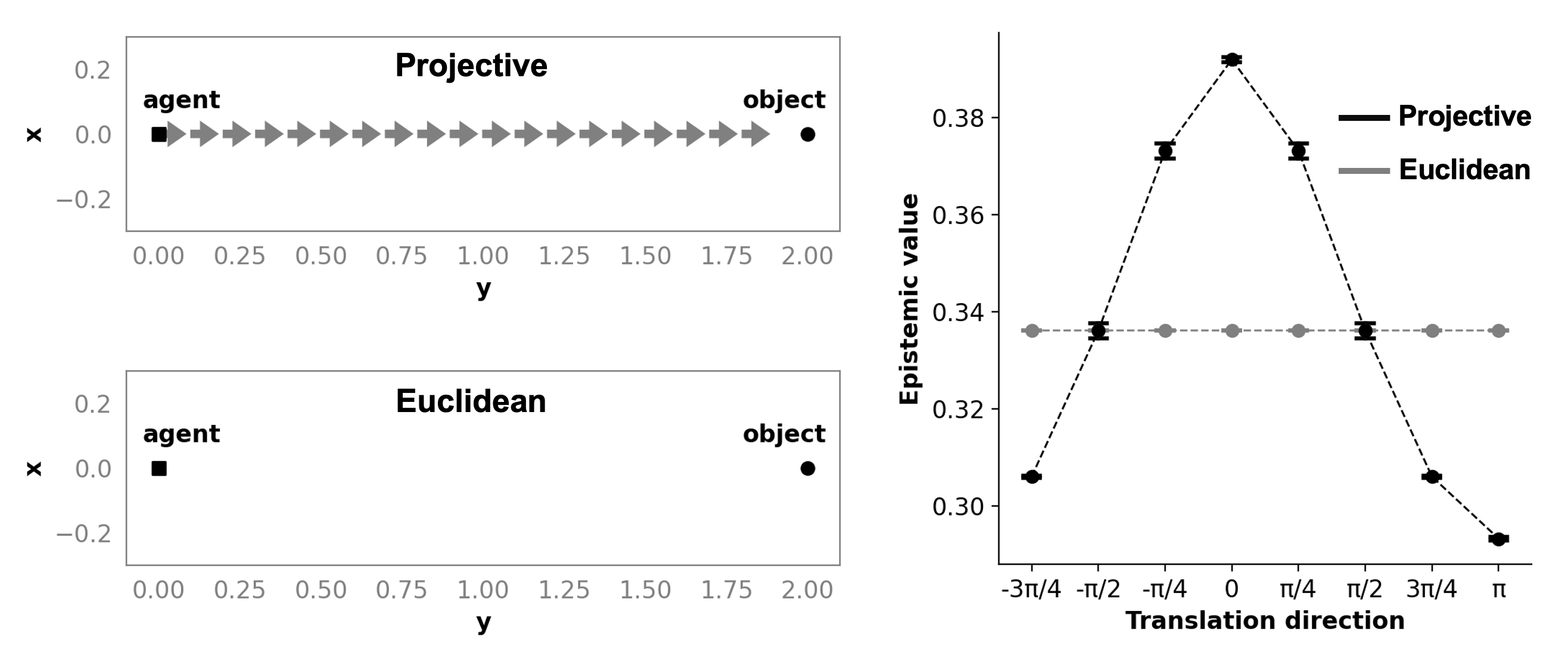}
    \caption{Simulation results}
    \medskip
    \small
    \raggedright
    {\textit{Left-tier.} Trajectory of the agent for the projective \textit{versus} Euclidean internal spaces. \textit{Right-tier.} Epistemic value as a function of directions of translation with respect to the object direction, for the projective \textit{versus} Euclidean internal spaces. Points are average values across comparable directions, and error-bars are standard errors.}
    \label{fig:simus}
\end{figure}

Figure \ref{fig:simus} left tier shows a representative example of trajectories obtained in the Euclidean \textit{versus} projective cases. In the projective case, the agent always approached the object. In the Euclidean case, the agent always stayed idle. Figure \ref{fig:simus} right tier shows epistemic value as a function of translation direction expressed in radians for both the projective and Euclidean cases. The direction of $0$ radian corresponds to the object direction. We averaged epistemic values across object positions within comparable directions. In the Projective case, average epistemic value was maximal for the direction of the object, and decreased with directions farther away from it. In the Euclidean case, average epistemic value was identical across directions. Maximum average epistemic value was much higher for the projective case than for the Euclidean case.      

\section{Discussion}

In this article, we introduced a generative model for environment exploration based on a first-person perspective in which actions are encoded as changes in perspective. The families of geometry for the world model encode possible `kinds' of perspective-taking on the environment and structures the representation of sensory evidence within the world model of the agent. In other words, each family corresponds to a specific perception and imagination scheme for the agent. We encoded two such families, namely the Euclidean \textit{versus} projective group as acting on the internal world model of the agent, i.e., within the geometric properties of this internal world model. We showed that different geometries induce different behaviors, focusing on the two cases: when the internal world model of the agent followed Euclidean geometry \textit{versus} projective geometry. This result contributes to understanding how integrative geometrical processing and principles can play a central role in cybernetics. In our approach, the geometry of the world model links perceptual and imaginary representations with actions and behaviors.

Although beyond the scope of this article, we are interested in generalizing the approach to compare the results obtained with those that could be obtained with other models in which geometry plays a central integrative role, e.g. as in \citep{ferreira2013bayesian} who used an internal 3-D egocentric, spherical representation of space. Such an approach would need to be expressed in terms of group action to make it formally comparable to our approach. Likewise, it would be interesting to compare other groups, and other more realistic models of sensors, and what types of behaviors they would induce. This would be useful to refine predictions and envision experimental designs for empirical validation in humans. We would also like to use more sophisticated settings (see for instance  \citealp{RUDRAUF2022110957,rudrauf2023combining}), even though it may be incompatible with the derivation of analytical solutions, but could lead to richer simulations and induction of behaviors.

We also wish to further examine how the geometry of a latent space intertwines with information processing. One motivation is theoretical, as we would like to assess how geometry changes learning behavior \citep{goyal2022inductive}. In this contribution, we have discarded representation learning \textit{per se}, as it was beyond the scope of this study focusing on planning. In future work, we intend to use deep learning to learn group structured representations. It is important to note that such approach differs from geometric deep learning \citep{https://doi.org/10.48550/arxiv.2104.13478,NEURIPS2022_474815da} as we do not seek to learn equivariant representations: a group structure will only be considered for the internal world model but none will be presupposed on the observation side.
Likewise, we are interested in examining how geometry may play a role in overt and covert attention.\\

Our contribution can simplify the design of novel agent architectures where exploratory and sensory choices of actions naturally emerge as a consequence of the internal representation and the reflection of the perceptual mechanism of the agent's embodiment. The study of world models with projective geometries was motivated by ongoing work in computational psychology aimed at reproducing features of consciousness. Projective geometry induces effects of magnification and focalization on information that appear immediately relevant for spatial attention, and more generally for contextual salience. Another motivation for this research is more practical, as we would like to use such principles to design virtual and robotic artificial agents mimicking human cognition and behaviors following  \citep{RUDRAUF2022110957,rudrauf2023combining}.

\bibliography{biblio-paper}

\appendix

\section{Proof of Proposition and Theorem}

\subsection{Proof of Proposition 1}

Any 3-D affine transformation is encoded by a matrix $M=(m_{i,j}; i,j=1..3)$ and a vector $(m_{j,4}; j=1..3)$; let $(m_{i,j}^{\mathcal{R}}; i,j=1..3)$ be the matrix associated to $\phi_{\mathcal{R}}$ and $(m_{4,j}^{\mathcal{R}}; j=1..3)$ its vector.

\underline{Projective case:}  $\phi^p_{\mathcal{R}}=\rho\circ\phi_{\mathcal{R}}$ is the projective map with expression in homogeneous coordinates given by the matrix, 

$$\begin{pmatrix}
m_{1,1}& m_{1,2} & m_{1,3}&m_{1,4}\\
m_{2,1}& m_{2,2} & m_{2,3}&m_{2,4}\\
m_{3,1}& m_{3,2} & m_{3,3}&m_{3,4}\\
0& 0 & \gamma &1 
\end{pmatrix}$$

By construction, the transition map in the projective case, $\psi_m^p$, is $\phi^p_{\mathcal{R}^m}\circ {\phi^p_{\mathcal{R}}}^{-1}$; it is the composition of two projective transformations, therefore it is a projective transformation.\\

\subsection{Proof of Theorem 1}\label{appendix-1}

 We will denote $B_y^{\epsilon}$ the Euclidean ball of radius 1 around $y\in \mathbb{R}^3$,\linebreak i.e. $B_y^{\epsilon}=\{ x\in \mathbb{R}^3 \vert \quad \Vert x-y\Vert \leq \epsilon \}$.
\begin{lem}
For any $Q\in \mathbb{P}(W)$, both in Euclidean and Projective cases, for any affine map or projective transformation $\psi:W\to W$,

\begin{equation}\label{important-equation}
C(\psi_{*}Q)=-\int dy  Q(\psi^{-1}( B_y^{\epsilon}))  \ln Q(\psi^{-1}( B_y^{\epsilon})) 
\end{equation}
\end{lem}

\begin{proof}

\begin{align}
C(\psi_{*}&Q)= \frac{3}{4\pi \epsilon^3}\times \nonumber \\
&\int \psi_{*} Q(d x_1) \int dy \1[x_1\in B_y^{\epsilon}] \ln \frac{\1[x_1\in B_y^{\epsilon}] }{\int  \psi_{*} Q(d x_1) \1[x_1\in B_y^{\epsilon}]} \nonumber 
\end{align}

\begin{align}
&=-\frac{3}{4\pi \epsilon^3}\int dy \ln Q(\psi^{-1}(B_y^{\epsilon})  \int \psi_{*} Q(d x_1)\1[x_1\in B_y^{\epsilon}] \nonumber\\
&= -\frac{3}{4\pi \epsilon^3}\int dy  Q(\psi^{-1}( B_y^{\epsilon}))  \ln Q(\psi^{-1}( B_y^{\epsilon})) 
\end{align}
\hfill\BlackBox

\end{proof}

\emph{Proof of Theorem:}\\

\underline{Euclidean case:}
for any set of moves $M$, and for any $m\in M$, $\psi_m$ is a rotation; therefore for any $y\in W$, $\psi^{-1}_m(B_y^{\epsilon})=B_{\psi^{-1}_m(y)}^{\epsilon}$. Then, for any prior $Q\in \mathbb{P}(W)$,

\begin{align}
C(\psi_{*,m}Q)&=- \frac{3}{4\pi \epsilon^3}\int dy Q(B_{\psi^{-1}_{m}(y)}^{\epsilon})\ln Q(B_{\psi^{-1}_{m}(y)}^{\epsilon}) \nonumber\\
&=-\frac{3}{4\pi \epsilon^3}\int dy Q(B_{y}^{\epsilon})\ln Q(B_{y}^{\epsilon})
\end{align}

In this case, the epistemic value is independent from the change of Euclidean frame, and not moving is a perfectly valid choice for the agent to maximize it, at each time step of the exploration algorithm (Algorithm 1).

\begin{rem}
The fact that staying still is a valid strategy arises as the agent assumes (or believes) that it has access to the whole configuration space of $O$. If it knew it had limited access to it, through for example limited sight, we expect the agent would look around until the object $O$ would be in sight, and then stop moving.
\end{rem}

\underline{Projective case:} 

Consider two projective transformations $\psi,\psi_1:W\to W$, if for any $y\in W$,

\begin{equation}
\psi^{-1}(B_y^{\epsilon})\subseteq \psi_1^{-1}(B_y^{\epsilon})
\end{equation}

then,

\begin{align}
-Q(\psi^{-1}(B_y^{\epsilon}))&\ln Q(\psi^{-1}(B_y^{\epsilon})) \noindent\\
&\geq -Q(\psi_1^{-1}(B_y^{\epsilon}))\ln Q(\psi_1^{-1}(B_y^{\epsilon}))
\end{align}

This suggests that the moves that maximize epistemic value are those where $\psi^{-1}_m$ shrinks the zone around $y^{o}=\rho(\phi_{\mathcal{R}}(o))$, which is the representation of $O$ in the internal world of the agent. In particular, it means magnifying the zone around $\rho(\phi_{\mathcal{R}^m}(o))$ in the agent's new frame, $\mathcal{R}^m$, after move $m$. The only way to do so is to select moves that bring the agent closer to $O$. Let us denote $y^{o}_m:= \rho(\phi_{\mathcal{R}^m}(o))$. Let us now make the previous argument more formal. We assume that the set of moves $M$ is finite. Let $Q_0=q_0 d\lambda$ be any initial prior on $W=P_3(\mathbb{R})$, at stating time $t=0$. 
After one step, move $m_1$ is chosen and the agent updates its prior as, 

\begin{equation}
q_1(x)\cong  1[x\in B_{y^{o}_{m_1}}^{\epsilon}]q_0(x)
\end{equation}

where $\cong $ means proportional to. The prior we now consider is $Q_1$ denoted simply as $Q$. One shows that there is $\alpha>0$, such that for all $m\in M$, and $\epsilon>0$ small enough,

\begin{align}
C(&\psi_{*,m}Q)= -\frac{3}{4\pi \epsilon^3}\times \nonumber\\
&\int dy \1[y \in B_{y^{o}_m}^{\alpha \epsilon}] Q(\psi^{-1}_m( B_y^{\epsilon}))  \ln Q(\psi^{-1}_m( B_y^{\epsilon})) \\
\end{align}

Let $\approx$ stand for `approximately equal to' (equal at first order in expansion in powers of $\epsilon$). Then from the previous statement the summand can be approximated by its value in $y^{o}_m$: 
\begin{equation}
C(\psi_{*,m}Q)\approx -\alpha^3 Q(\psi^{-1}_m( B_{y^{o}_m}^{\epsilon}))  \ln Q(\psi^{-1}_m( B_{y^{o}_m}^{\epsilon}))
\end{equation}

Furthermore, $Q(\psi^{-1}_m( B_{y^{o}_m}^{\epsilon})) \approx \frac{4\pi\epsilon^3}{3} \frac{q_1(y^{o})}{\vert \det \nabla \psi_m\vert (y^{o})}$, where $\vert \det \nabla \psi_m\vert (y^{o})$ is the absolute value of the Jacobian determinant of $\psi_m$ at $y^{o}$. The epistemic value is maximized when $\vert \det \nabla \psi_m\vert (y^{o})$ is maximized. By definition, $\psi_m= \rho \circ \phi_{\mathcal{R}^m}\circ \phi_{\mathcal{R}}^{-1}\circ\rho^{-1}$, therefore, by the chain rule of 
differentiation 

\begin{align}
 &\vert\det \nabla \psi_m\vert (y^{o})  \nonumber\\
 &= \vert\det \nabla \rho\vert (\phi_{\mathcal{R}^m}(o)). \vert \det \phi_{\mathcal{R}^m}\vert(o). \vert \det \nabla [\phi_{\mathcal{R}}^{-1}\circ\rho^{-1}]\vert(y^{o})
\end{align}

Let us make explicit each terms in the previous equation. 
$\phi_{\mathcal{R}^m}$ is a rigid movement therefore, $\vert \det \phi_{\mathcal{R}^m}\vert(o)=1$. $\vert \det \nabla [\phi_{\mathcal{R}}^{-1}\circ\rho^{-1}]\vert(y^{o})$ does not depend on $m$ so we can label it as a constant $C$. $\phi_{\mathcal{R}^m}(o)$ is the coordinate of $O$ in the Euclidean frame $\mathcal{R}^m$; let us denote $(x^m,y^m,z^m)$ these coordinates, i.e. $(x^m,y^m,z^m):=\phi_{\mathcal{R}^m}(o)$. Then,

\begin{equation}
\vert\det \nabla \rho\vert (x^m,y^m,z^m)= \frac{1}{(\gamma z^{m}+1)^4}
\end{equation}

Therefore, $\vert \det \nabla \psi_m\vert (y^{o})= C\frac{1}{(\gamma z^{m}+1)^4}$.

As we assumed that for any move $m\in M$, the object $O$ is always in front of the agent, then $z^{m}\geq 0$; in this case, $z^{m}$ is also the distance of the agent to the object. Epistemic value is maximized when $z^{m}$ is minimized and therefore the agent selects moves that reduce its distance to the object. Denote one of such move $m^*$; the argument then loops back with the new reference frame $\mathcal{R}^{m^*}$ and updated belief $q\gets\psi_{m,*} q_{\vert y^{o}_m}$.
\hfill\BlackBox

\end{document}